\documentclass[letterpaper, 9 pt, conference]{ieeeconf}  % for US Letter paper

\IEEEoverridecommandlockouts
\usepackage{cite}
\usepackage{amsmath,amssymb,amsfonts}
\usepackage{algorithmic}
\usepackage{graphicx}
\usepackage{textcomp}
\usepackage{xcolor}
\usepackage{gensymb}
\usepackage{caption}
\usepackage{subcaption}
\usepackage{hyperref}
\usepackage{algorithm}

\usepackage{array}

\usepackage{listings}

\usepackage{booktabs}
\usepackage{multirow}
\usepackage{siunitx}

\def\BibTeX{{\rm B\kern-.05em{\sc i\kern-.025em b}\kern-.08em
    T\kern-.1667em\lower.7ex\hbox{E}\kern-.125emX}}

\title{Interleaved LLM and Motion Planning for Generalized Multi-Object Collection in Large Scene Graphs}

\author{Ruochu Yang$^{1}$, Yu Zhou$^{2}$, Fumin Zhang$^{3}$, Mengxue Hou$^{2}$
\thanks{This research work is supported by ONR grants N00014-19-1-2556 and N00014-19-1-2266;  AFOSR grant FA9550-19-1-0283; NSF grants GCR-1934836, CNS-2016582 and ITE-2137798; and NOAA grant NA16NOS0120028.}
\thanks{$^{1}$School of Electrical and Computer Engineering, Georgia Institute of Technology, Atlanta, USA. $^{2}$School of Electrical Engineering, Notre Dame University, USA. $^{3}$Department of Electrical and Computer Engineering, Department of Mechanical and Aerospace Engineering, The Hong Kong University of Science and Technology, Hong Kong, China.}
}

\begin{document}

\maketitle
\thispagestyle{empty}
\pagestyle{empty}

\begin{abstract}

Household robots have been a longstanding research topic, but they still lack human-like intelligence, particularly in manipulating open-set objects and navigating large environments efficiently and accurately. To push this boundary, we consider a generalized multi-object collection problem in large scene graphs, where the robot needs to pick up and place multiple objects across multiple locations in a long mission of multiple human commands. This problem is extremely challenging since it requires long-horizon planning in  a vast action-state space under high uncertainties. To this end, we propose a novel interleaved LLM and motion planning algorithm \textit{Inter-LLM}. By designing a multimodal action cost similarity function, our algorithm can both reflect the history and look into the future to optimize plans, striking a good balance of quality and efficiency. Simulation experiments demonstrate that compared with latest works, our algorithm improves the overall mission performance by 30\% in terms of fulfilling human commands, maximizing mission success rates, and minimizing mission costs.

\end{abstract}

\section{Introduction}
\label{intro}

\begin{figure}[ht]
        \centerline{\includegraphics[width=0.5\textwidth]{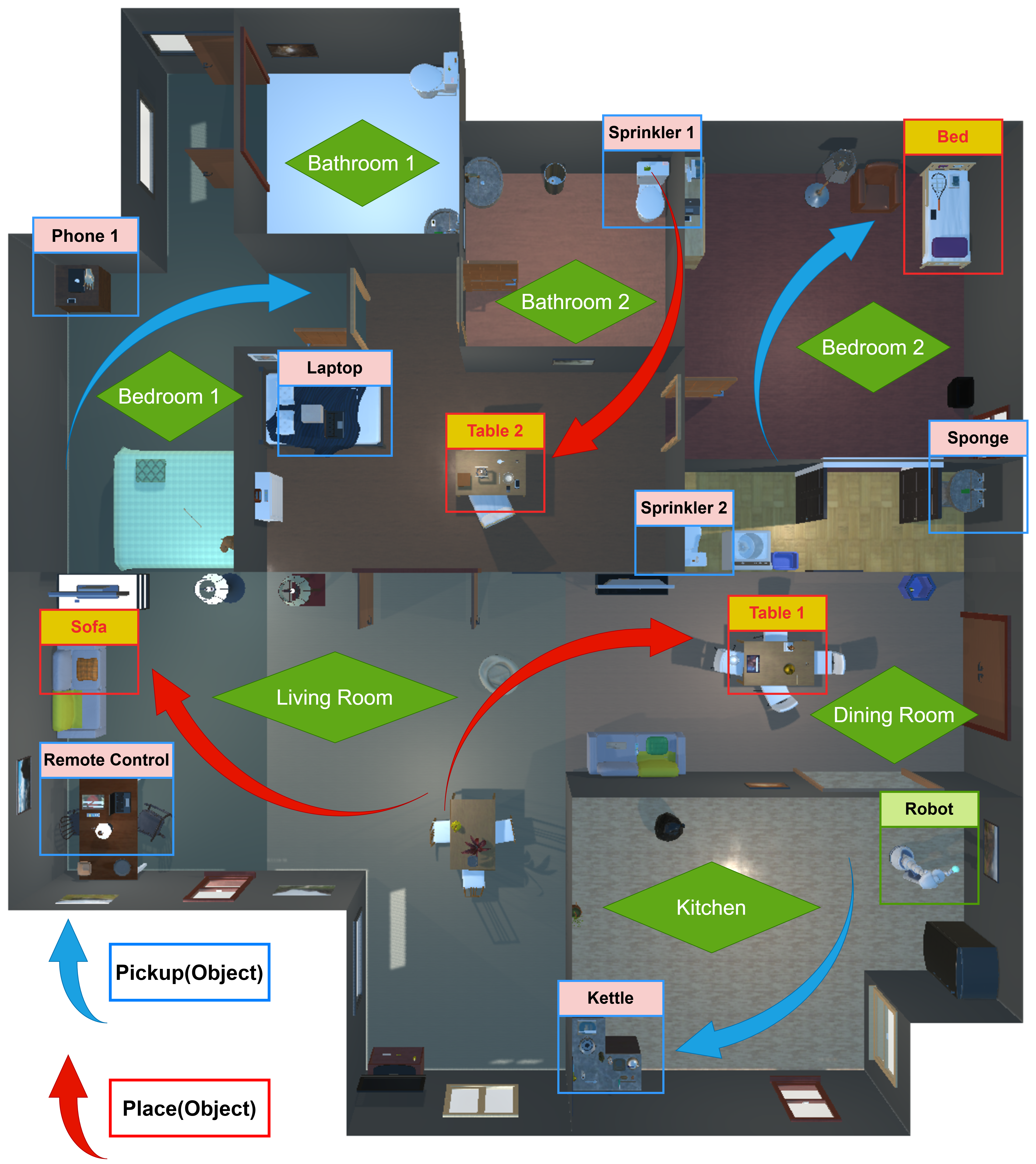}}
        \caption{Generalized Multi-Object Collection in Large Scene Graphs. In a complicated large environment with many rooms/furnitures/objects, the robot needs to plan navigation and manipulation actions to accomplish a long-horizon mission consisting of an open set of human commands.}
        \label{figure introduction}
\end{figure}

%Good reference: Online Planning for Large Markov Decision Processes with Hierarchical Decomposition

%% motivation (what do people care about in this field, what do I want to do) 
The robotics community has been aiming to develop household robots \cite{gervet2023navigating, wu2023tidybot}, where task and motion planning (TAMP) \cite{garrett2021integrated} is an indispensable research domain. However, many works have focused on compact and deterministic environments \cite{jiao2022sequential, zhu2021hierarchical}, struggling to scale to complex real-world environments with open-vocabulary objects, long mission horizons, and unstructured workspaces \cite{ray2024task}. Many works \cite{rana2023sayplan, rajvanshi2024saynav, singh2023progprompt} have explored the semantic reasoning capability of Large Language Models (LLMs) for household tasks. However, these works primarily focus on high-level semantic planning based on LLM commonsense heuristics while neglecting real-world execution costs or physical constraints \cite{valmeekam2022large}, making their generated plans inefficient or even infeasible for robotic execution. Therefore, we are motivated to answer this question ``For household robots to fulfill generalized human needs, how can we devise a scalable planning algorithm that achieves a balance of efficiency and quality?''

%% problem formulation (a specific task in a specific scenario)
There has been a trend of endowing human-like intelligence with household robots by defining more and more complicated tasks \cite{yenamandra2023homerobot, honerkamp2024language}. Following this trend, we imagine the future for household robots should be towards multi-user, multi-modal, and multi-scenario, i.e., long-horizon missions in generalized environments. Imagine a family get up in a busy morning and each family member issues commands to a household robot for help. For example, mother would like to have the breakfast while father is having an online meeting in 10 minutes. To this end, we consider a difficult yet unresolved task \textit{generalized multi-object collection} as shown in Figure \ref{figure introduction}. The term \textit{generalized} highlights a long mission with multiple human commands issued by multiple users in a short time window. To fulfill these commands, a robot needs to reason about object-place semantics and move desired objects from place to place in a large environment accurately and efficiently. During this long-horizon process, the robot may face challenges in the realistic world like narrow pathways or hard-to-reach objects. All the above factors constitute a very hard planning problem.

%% contribution (propose our method to target the problem, show the novelty of our method)
To the best of our knowledge, this is the first time that an interleaved algorithm of LLM and motion planning is proposed to tackle this hard problem. While the robot is fulfilling a long-horizon mission, these two planners consistently share information with each other. Our key insight is that through consistent feedback from motion planning, the LLM planner can be grounded with more accurate cost functions to hedge against a number of uncertainties in the large environment. Specifically, we design a multimodal action cost similarity function which can both reflect the history and look into the future to systematically optimize the cost functions. Eventually, our interleaved algorithm will generate near-optimal plans which minimize mission costs and maximize success rates. Our main contributions are summarized as follows:
\begin{itemize}
    \item Pushing towards human-like household robots, we consider a complicated \textit{generalized multi-object collection} problem in large scene graphs. To solve this challenging problem, we propose a novel interleaved LLM and motion planning algorithm \textit{Inter-LLM}. Our key design is a multimodal similarity function which estimates unknown action costs for LLM plan pruning, thus approaching near-optimal mission performance on the fly. 
    
    \item We evaluate our proposed algorithm on a household robot (embodied as a mobile manipulator) in a photo-realistic simulator featuring physical interactions with the environment. In a long-horizon mission consisting of multiple abstract human commands, the robot can iteratively achieve better performance with faster planning speed. Through baseline comparison, our algorithm outperforms latest scene graph planning works.

\end{itemize}

\section{Related Works}
\label{related works}

\subsection{Task and Motion Planning}

Since TAMP is capable of solving hierarchical semantic/geometric problems \cite{zhu2021hierarchical}, the robotics community recently explores its potential in scene graph planning \cite{jiao2022sequential}. A latest work \cite{ray2024task} theoretically formulates TAMP in scene graphs, but it focuses on navigation without object manipulation. Moreover, it relies on sequential planning, while we leverage motion costs in an interleaved manner. \cite{yang2024oceanplan} proposes a three-layer LLM-task-motion planning method for underwater vehicles. However, the three planners operate sequentially without any internal feedback for optimizing the whole plan.

TAMP has three planning workflows: sequence-first, satisfaction-first, and interleaved. Sequence-first workflows \cite{wei2022chain, shen2024hugginggpt, lu2024chameleon} rely on LLMs to pre-determine a full plan and then execute it in an open-loop manner. While efficient, this workflow overlooks optimal solutions in the big decision space of large environments and long missions. Satisfaction-first workflows \cite{hao2023reasoning, yao2024tree} generate multiple solutions through exhaustive sampling, then identify the best one by LLMs. However, it becomes computationally infeasible in large scene graphs with vast action-state space. Our work is motivated by the interleaved workflow of \cite{hou2023interleaved} which balances planning efficiency and quality by estimating motion costs to prune task planning branches. But their focus is theoretical, lacking practical considerations for long missions in large environments. Moreover, their task planning is formulated in the symbolic space which can't process rich semantics in scene graphs.

\subsection{LLM Planning}
Recently LLMs have become a powerful way of commonsense reasoning towards open-vocabulary scenarios. \cite{liu2023llm} use LLMs to generate high-level PDDL files for a classical task planner to solve. \cite{qin2023toolllm} develops a LLM-based orchestrator to master 16000+ APIs with execution details hidden in the APIs. \cite{zhuang2023toolchain, hao2023reasoning} go one step further by considering costs of high-level APIs but computing it from LLM heuristics (lexically close words). \cite{rajvanshi2024saynav} employs LLM planning to explore new environments but limits to sequential search and lacks low-level action feedback. \cite{rana2023sayplan} utilizes LLMs for iterative replanning on sub-graphs but neglects costs associated with motion planning by treating it as a blackblox executor. However, our interleaved algorithm constantly collects real-world execution costs to make our plan more cost-efficient.

\subsection{Robotic Planning in Scene Graphs}
In recent years, scene graph planning for household robots emerge as a representative domain. Early studies \cite{zhou2023esc, schmalstieg2022learning} focus on the basic problem of static object search, while \cite{schmalstieg2023learning} further considers interacting with the environment to explore more hidden objects. Later \cite{zheng2023asystem} extends to generalized multi-object search in 3D environments. Most recently, \cite{yenamandra2023homerobot} proposes an open-vocabulary mobile manipulation challenge to facilitate everyday household tasks, and \cite{honerkamp2024language} extends an interactive semantic search task from their previous work \cite{schmalstieg2023learning} for more complex scenarios. Motivated by these works, we focus on a challenging yet unresolved problem of generalized multi-object collection.

Conceptually most similar to our work, \cite{honerkamp2024language} grounds an LLM planner in dynamically updated scene graphs. However, they focus primarily on high-level LLM reasoning, neglecting low-level action costs. For example, they assign the same cost to all \textit{open/close} actions, regardless of differences between objects and workspaces. We aim to bridge this gap by proposing an interleaved algorithm to explicitly consider costs of object grasping and room traversal, thus enabling the robot to learn from real-world physics rather than solely rely on LLM semantic reasoning.

\section{Problem Formulation}
\label{problem formulation}

We consider a challenging scene graph planning problem \textit{generalized multi-object collection} which requires long-horizon planning and extensive space navigation. Situated in a large household environment of objects/furnitures/rooms, a robot needs to fulfill a long-horizon mission $\mathcal{Q} $ consisting of multiple human commands $\{q_1,...,q_N\}$. Each command can be delivering multiple objects to multiple locations.

% Specifically for household robots, there are two main types of planning - navigation and manipulation. We jointly consider their interdependent effects for this comprehensive yet complicated scene graph planning problem. For example, a task plan can be a long sequence of actions \textit{navigate(location) - pickup(object) - navigate(location) - place(object) - navigate(location)-...}, where the selection of \textit{navigate} action and its \textit{location} parameter will influence the next \textit{pickup} action and its \textit{object} parameter, and vice versa. 

% To fulfill these complex human commands $\mathcal{Q}$, the robot has to perceive and interact with the environment while planning a long sequence of actions through exploration/interaction feedback.
%\cite{honerkamp2024language} say "if trying to open objects that cannot be opened, the according subpolicy will fail and the LLM has to reason about an appropriate response". Can LLM really give a corrected plan? Instead, can we use the real motion cost in the physical world? Will this benefit the planning?  

\subsection{Graph Representations of Household Environment}

To represent the household environment where the robot plans and acts, we formulate two types of graphs as follows.

\textbf{Scene Graph $\mathcal{G}^s$:} 
We model the environment with a pre-built scene graph which holds rich semantic information at different abstraction levels $\mathcal{G}^s = \langle \mathcal{N}^s, \mathcal{E}^s \rangle $. The nodes $\mathcal{N}^s$ consist of $N_o$ objects $Obj = \{ obj_1, ..., obj_{N_o} \} $,  $N_f$ furnitures $Fur = \{ fur_1, ..., fur_{N_f} \}$, and $ N_r$ rooms  $Rm= \{rm_1,...,rm_{N_r} \} $. The nodes also denote semantic attributes such as usage like \textit{cup for drinking beverage}. The edges $\mathcal{E}^s$ denote spatial relations between the objects/furniture/rooms like \textit{cup on the table}.

% The scene graph is organized in a hierarchical manner with three primary levels: rooms, furniture, and objects. These rooms are interconnected through pose nodes to represent the environment’s topological structure. Within each room, we find objects (movable or operable entities).  Object nodes encode particulars including state, affordances, additional attributes such as color or weight, and 3D pose.

%We model the large-scale multimodal environment as a hierarchical scene graph $\mathcal{G}^s = <V,E> $, where the set of nodes $V = V_1 \bigcup V_2 \bigcup ... \bigcup V_K $ represents the set of vertices at a particular level of the hierarchy $i$. Edges $E$ stemming from a vertex $v \in V_i, i=1,...,K$ may only terminate in $V_{i-1} \bigcup V_i \bigcup  V_{i+1} $, i.e. edges connect nodes within the same level, or one level higher or lower.

\textbf{Occupancy Grid Graph $\mathcal{G}^o$:}
In addition to the scene graph $\mathcal{G}^s$, we define an occupancy grid graph to represent local workspace geometry $\mathcal{G}^o = \{\mathcal{N}^o\}$, where each node $n^o \in \mathcal{N}^o $ denotes a 2D location as occupied or free. It models a local area occupied by a room or furniture with a clear contour, but no exact locations of the objects. This is a widely-accepted assumption \cite{rosinol20203d} since 1) it is computationally heavy to maintain such a high-resolution graph with exact locations of many (tiny) objects; 2) objects tend to be moved from place to place during long-horizon missions.

% \textbf{Topological Graph $\mathcal{G}^p$:} In addition to the semantic scene graph, we hold a topological graph $\mathcal{G}^p = \langle \mathcal{N}^p, \mathcal{E}^p,  \mathcal{W}^p \rangle $ for grounded navigation across the large-scale environment. The nodes $\mathcal{N}^p$ are reachable 2D locations for the robot, the edges $\mathcal{E}^p$ connect these nodes, and the edge weights $\mathcal{W}^p$ represent navigation costs of moving from one location to another. This topological graph serves as a way of representing geometric details of the workspace where the robot is physically acting.

\subsection{Hierarchical Formulations for Planning}

As we present above, this scene graph planning problem is difficult to solve, underlaid by a large action-state space and a long task horizon. Motivated by TAMP\cite{garrett2021integrated}, we convert the original problem into an hierarchical one to decompose planning complexity.

% We formally model the planning problem with a Markov Decision Process (MDP) $\mathcal{M} = \langle \mathcal{S}, \mathcal{A}, \mathcal{O}, T, R  \rangle$, where  $\mathcal{S}, \mathcal{A}, \mathcal{O}$ denote the state space, action space, and observation space, and $T, R$ denote the transition model and reward model. We explain the MDP formulation in details as follows.

% In our specific problem formulation, an action $a(\delta) \in A$ is modeled by a function $a$ such as \textit{pickup}, a semantic parameter $\delta$ such as an object to pick up. After instantiated by a semantic parameter like \textit{cup}, each action like \textit{pickup(cup)} is fulfilled for real by an underlying control policy $\pi$ with a geometric parameter $\tau$ such as a manipulation trajectory.

\textbf{High-level State Space $\mathcal{S}^h$:}
We define the high-level state space $\mathcal{S}^h$ as follows, which reflects the robot's physical state in conjunction with the scene graph nodes $\mathcal{N}^s$. 
\begin{itemize}
    \item  \textit{holding(object)}: the object which the robot is grasping.

    \item \textit{hand\_free}: if no object is grasped by the robot or not.

    \item \textit{at(furniture)}: the furniture at which the robot is.
    
     \item \textit{at(room)}: the room in which the robot is.
\end{itemize}

\textbf{Low-level State Space $\mathcal{S}^l$:}
The low-level state space $\mathcal{S}^l$ is defined as the robot's reachable 2D locations, i.e., the free nodes at the occupancy grid graph $\mathcal{G}^o$.

\textbf{High-level Action Space  $\mathcal{A}^h$:}
We denote the high-level action space $\mathcal{A}^h$ with each high-level action $a_k \in \mathcal{A}^h$ meant for high-level LLM planning. We define a set of API functions $\mathcal{F} = \{f_0, f_1, ..., f_m \}$, where each API function is associated with a clear text description. Specifically, $a_k \in \mathcal{A}^h$ is defined in terms of an API function $f \in \mathcal{F}$ with a node $ n^s \in \mathcal{N}^s$ in the scene graph, i.e.,  $a_k  \triangleq \langle f, n^s \rangle, \mathcal{A}^h  \triangleq \mathcal{F} \times \mathcal{N}^s $. We define $\mathcal{A}^h$ as follows: 1) \textit{navigate(furniture, room)} - navigate to a furniture in a room; 2) \textit{pickup(object, furniture)} - pick up an object on a furniture; 3) \textit{place(object, furniture)} - place an object on a furniture.

\textbf{Low-level Control Space $\mathcal{A}^l$:}
Given $a_k \in \mathcal{A}^h$, the low-level motion planner should try to fulfill it in the real world. This process is driven by an inherent control policy $\pi_{a_k}$ which generates reasonable control inputs $u_t$ in the low-level control space $\mathcal{A}^l$. We define $u_t \in \mathcal{A}^l$ as follows: 1) move forward by $5cm$; 2) turn left by $45 \degree$; 3) turn right by $45 \degree$; 4) pick up an object at a 2D location; 5) place an object at a 2D location.

\subsection{Explicit Cost Function of Actions}

Executing actions inherently incurs costs based on robot dynamics and local workspaces, like picking up a tiny cup from a cluttered table. Our key innovation lies in explicitly incorporating these action costs for globally optimal planning. For executing $a_k \in \mathcal{A}^h $ at a high-level state $s^h_k \in \mathcal{S}^h $, we explicitly define its associated cost function $c(a_k, s^h_k)$. Embodied for household robot dynamics, the cost function can be instantiated into the following categories.

\begin{itemize}
    \item Navigation Cost $c^{nav}$: when $a_k$ is \textit{navigate(room/furniture)} and $s^h_k$ is \textit{hand\_free}, the cost function $c(a_k, s^h_k)$ is instantiated as $c^{nav}(a_k, s^h_k)$.

    \item Object Manipulation Cost $c^{man}$: when $a_k$ is \textit{pickup(object)} or \textit{place(object)} and $s^h_k$ is \textit{at(furniture)}, the cost function $c(a_k, s^h_k)$ is instantiated as $c^{man}(a_k, s^h_k)$.
\end{itemize}

%\textbf{Object Carrying Payload $c^p$:} When $a_k$ is \textit{navigate(room/furniture)} and $s^h_k$ is \textit{holding(object)}, the cost function $c(a_k, s^h_k)$ is instantiated as a object carrying payload $c^p(a_k, s^h_k)$. This cost category denotes the payload of robot carrying an object to a destination location.

%\textbf{Object Instance Ambiguities} Ambiguities of multiple instances of the specified class in a room are resolved by selecting the closest instance. The subpolicies then generate actions in the low-level action space and return once they succeed or encounter a failure. Throughout their execution, they continuously update the scene representations.

\subsection{Overall Planning Problem}

Based on the above concrete definitions, we present our overall planning problem of \textit{generalized multi-object collection} in large scene graphs. Given an initial high-level state $s_{init}$ and a long-horizon mission $\mathcal{Q}$, our goal is to obtain an optimal set of plans $ \{ \Pi_{q_1},..., \Pi_{q_N} \}^* $, which minimize the total costs of accomplishing all the human commands $ \{ q_1,...,q_N \} \in \mathcal{Q} $ as follows:
\begin{gather}
\label{equation overall problem}
     \{\Pi_{q_1},..., \Pi_{q_N}\}^* =  \arg \min_{\{\Pi_{q_1},..., \Pi_{q_N} \}}   J_{total}  \\
       J_{total} =  \sum_{i=1}^N J_{q_i}   = \sum_{i=1}^N \sum_{k=1}^T c(a_k, s^h_k) \\
    \text{s.t.  }   \Pi_{q_i} = \{ a_k \}_{k=1}^T,  a_k \in \mathcal{A}^h \\
   s^h_{k+1}  = \mathcal{V}(a_k, s^h_k) \\
    s^h_0 = s_{init}, s^h_k = s_{goal}, 
\end{gather}
where $c(\cdot, \cdot)$ is the corresponding action cost,  $\mathcal{V}(\cdot, \cdot)$ is the underlying environment dynamics, and $s_{goal}$ is a goal high-level state interpreted from each command $q_i \in \mathcal{Q}$.

It is extremely complicated (NP-hard) to solve the overall planning problem \eqref{equation overall problem}, i.e., finding all the optimal plans $ \{ \Pi_{q_1},..., \Pi_{q_N} \}^*$ with minimum total costs $J^*_{total}$. Essentially, solving a planning problem is determined by two factors of the search tree: branches (action-state pairs) and depth (task temporal horizon). Specific to our problem \eqref{equation overall problem}, the branch factor is underlaid by rooms/furniture/objects in the scene graph $\mathcal{G}^s$, and the depth factor is underlaid by the long mission $\mathcal{Q}$. Therefore, there always exits a tradeoff of thoroughly exploring the search tree while efficiently pruning high-cost branches.

%%%%%%%%%%%%%%%%%%%%%%%%%%%%%%%%%%%%%%%%%%%%%%%%%%%
\section{Methodology}
\label{methodology}

%\begin{figure*}[ht]
   %\centerline{\includegraphics[width=\textwidth]{method/framework.png}}
    %    \caption{Interleaved LLM and Motion Planning.}
     %   \label{figure framework}
% \end{figure*}

To solve the challenging problem \eqref{equation overall problem}, we propose a novel algorithm of interleaved LLM and motion planning named \textit{Inter-LLM}. The designs of our algorithm are two-fold. First, it is a hierarchical algorithm which can decompose the complex problem into multiple smaller ones. After the high-level LLM planner fixes the task plan, the low-level motion planner can solve it in significantly reduced search space. Second, our algorithm implements interleaved interaction between the two-level planners. Estimated costs derived by the motion planner help rule out branches unlikely to be the optimal solution for the LLM planner, thus achieving a good balance of quality and efficiency. The main algorithm is presented in Algorithm \ref{main algo}.

\begin{algorithm}
\caption{Main Algorithm of Interleaved LLM and Motion Planning}
\label{main algo}
\begin{algorithmic}[1]
\REQUIRE  Long-horizon mission $\mathcal{Q}$ consisting of $N$ human commands $ \{q_1,...,q_N\}$, Semantic scene graph $\mathcal{G}^s$, Occupancy grid graph $\mathcal{G}^o$, High-level action space $\mathcal{A}^h$, High-level state space $\mathcal{S}^h$, Underlying environment dynamics $\mathcal{V}$, Feasibility checker $\mathcal{F}^c$, Multimodal similarity function $\mathcal{F}^{ms}$

\FOR{$q_i \in Q, i=1,...,N$} 

    \STATE High-level LLM planner generates $M$ task plan candidates $\{\Pi^{q_i}_1, ..., \Pi^{q_i}_M \}$ based on $\mathcal{G}^s$

    \FOR{$j = 1, ... ,M$}

        \WHILE{True}
            \STATE Check feasibility $ valid/invalid \Leftarrow \mathcal{F}^c(\Pi^{q_i}_j)$
            
            \IF{$invalid$}
               \STATE  LLM re-generates a plan candidate $ \Pi^{q_i}_j$
            \ENDIF

        \ENDWHILE
        
    \ENDFOR
    
    \STATE Estimate the total cost of each task plan candidate $\hat{c}^{\Pi^{q_i}_j}_{total} \Leftarrow  \mathcal{F}^{ms}(\Pi^{q_i}_j), j=1,...,M$
   
    \STATE Select the best task plan $\Pi_*^{q_i} = \arg\min_{\hat{c}^{\Pi^{q_i}_j}_{total}}\Pi^{q_i}_j, j=1,...,M$, where $\Pi_*^{q_i} = \{a_1,...,a_T\}$
    
    \FOR{$a_k \in \Pi_*^{q_i}, k=1,...,T$} 
    
        \STATE  Low-level motion planner executes the action $a_k$ at the current high-level state $s^h_k$ in the real world 
        
        \STATE  Sample the empirical action cost $\tilde{c}(a_k, s^h_k)$ based on $\mathcal{G}^o$

        \STATE  Update known action costs  $C^{known} \Leftarrow \tilde{c}(a_k, s^h_k)$

        \STATE  Update state $s^h_{k+1} = \mathcal{V}(a_k, s^h_k)$
          
    \ENDFOR
    
\ENDFOR
\end{algorithmic}
\end{algorithm}

\subsection{High-level Graph Search LLM Planner}

% {\color{red} Prime the Search: Using Large Language Models for Guiding Geometric Task and Motion Planning by Warm-starting Tree Search}

Given our problem \eqref{equation overall problem}, the LLM needs to plan towards the scene graph $\mathcal{G}^s$. Therefore, we formulate the task planning process as a LLM-based graph search. We set the root node of the search graph as the current high-level state $s^h_k$ and each branch as a candidate high-level action $a_k \in \mathcal{A}^h$ for selection. We design the following steps to implement the LLM graph search for generating accurate and valid task plans. Specifically, the LLM needs to 1) reason about environment semantics to fulfill multiple human commands; 2) plan a long sequence of actions instantiated by its parameter; 3) incorporate action costs from the motion planner to filter out unpromising actions.

\textbf{Prompt Design:}
The LLM prompt consists of the current human command $q_t$, the current high-level state $s^h_k$, the current scene graph $\mathcal{G}^s_t$, and the high-level actions $\mathcal{A}^h$ with clear text descriptions. Additionally, each action must be fulfilled with an entity parameter, i.e., room/furniture/object node $\mathcal{N}^s \in \mathcal{G}_t^s$. This is essentially using LLM semantic heuristics to search in the scene graph to concretize the high-level action candidate. Since we need LLM to concretize actions by considering a set of rooms/furnitures/objects, we provide LLM with the full JSON-formatted scene graph. Since in our interleaved algorithm (we will introduce it later), we ask LLM to consistently incorporate cost feedback from motion planner to prune high cost actions when generating a task plan. We ask the LLM to generate $M$ number of task plan candidates and make sure each task plan is different from each other.

% Essentially, our semantic tree search is a heuristic best-first search algorithm.

\begin{figure}
\centerline{\includegraphics[width=0.5\textwidth]{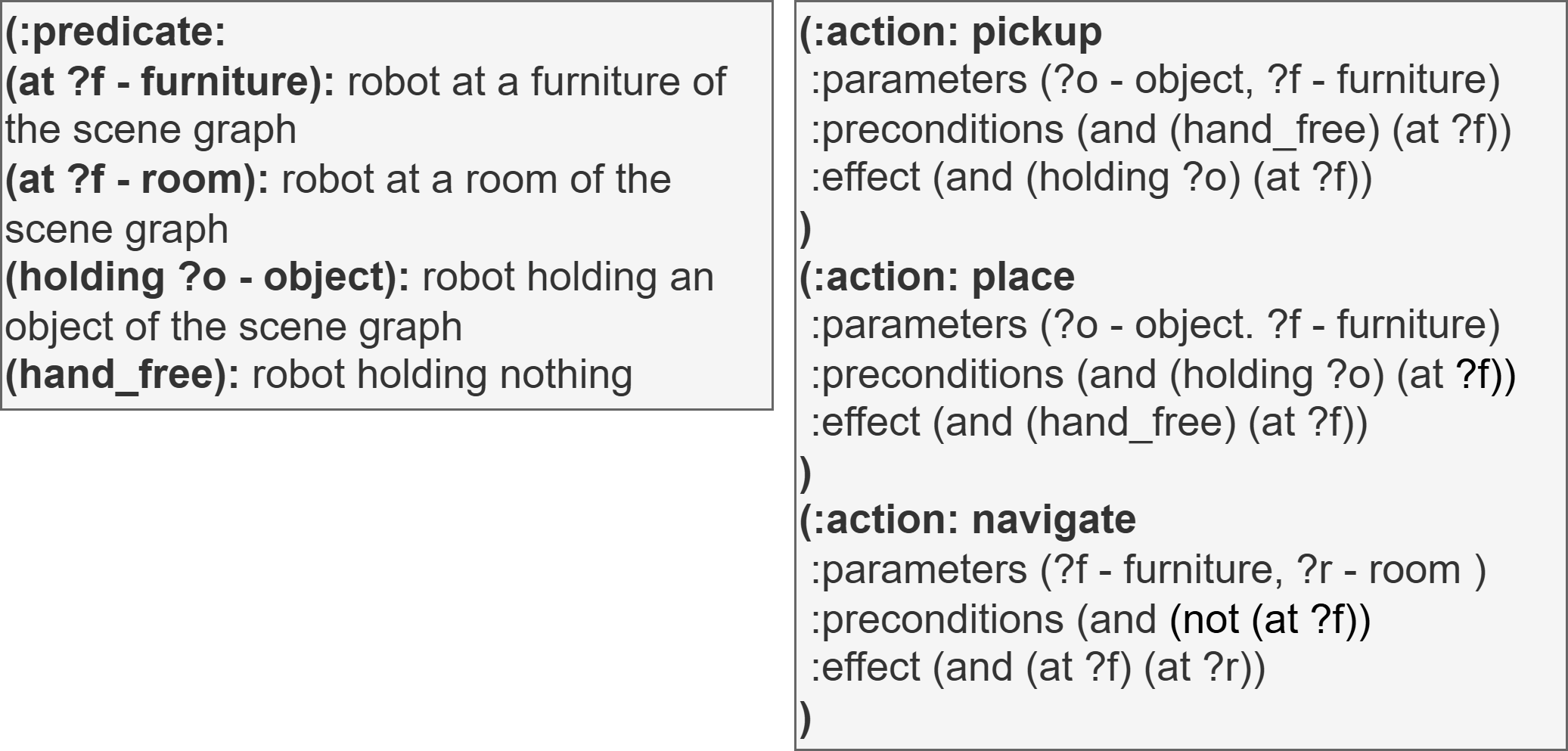}}
        \caption{Predicates, preconditions, and effects of high-level actions in a task plan.}
        \label{figure pddl}
\end{figure}

\textbf{Feasibility Checker $\mathcal{F}^c$:}
Since we provide LLM with the full JSON-formatted scene graph, LLM hallucinations are likely to happen \cite{honerkamp2024language}. This is exacerbated when LLM plans over large environments or long missions. Therefore, we propose a feasibility checker $\mathcal{F}^c$ to correct invalid actions in the LLM-generated task plan. First, we check if the high-level action itself $a \in \mathcal{A}^h$ is feasible, i.e., follows the rule of preconditions and effects. Correspondingly, we define the action predicates based on the high-level state space $\mathcal{S}^h$. As shown in Figure \ref{figure pddl}, we design the logical rule by following a standard PDDL paradigm \cite{aeronautiques1998pddl} and then check if each action follows the rule.
Second, we follow the below rules to check if each parameter of the action, i.e., the node $ n^s \in \mathcal{N}^s$, is feasible regarding the scene graph $\mathcal{G}^s$.
\begin{itemize}
    \item  object not in the scene graph $obj_i \notin \mathcal{G}^s $
    \item  furniture not in the scene graph $ fur_i \notin \mathcal{G}^s$
    \item  room not in the scene graph $ rm_i \notin \mathcal{G}^s$
    \item  $obj_i$ is picked from $fur\_i$ but robot navigated to $fur_j$
    \item  $obj_i$ is placed on $fur\_i$ but robot navigated to $fur_j$
    \item  $obj_i$ is neither picked up nor in hand
\end{itemize}
Finally, any feasibility violation will be updated into the LLM planning prompt and the LLM re-generates a task plan until feasible.

\subsection{Low-level Sampling-based Motion Planner}

Since the LLM planner is unaware of action costs in the high-dimensional continuous space, the motion planner is key to grounding it with real-world physics. Specifically, the high-level action $a_k$ planned by LLM is passed to the motion planner for real-world execution at the current high-level state $s^h_k$. During the execution process, the motion planner tries to obtain the action cost in a local workspace. Usually, the true action cost $c(a_k, s^h_k)$  is unknown and it cannot be analytically derived because of high-dimensional action dynamics (7 DoF manipulation) and unmodeled local workspace (cluttered table). Therefore, the best we can do is resorting to sampling-based methods to obtain an empirical cost $\tilde{c}(a_k, s^h_k)$. As future work, we can use RL to obtain the action cost by offline training from empirical trials \cite{brohan2023can}. At the current high-level state $s^h_k$, we uniformly sample $N_l$ low-level states $s^l_1, ..., s^l_{N_l} \in \mathcal{S}^l$, i.e., free nodes in the occupancy grid graph $\mathcal{G}^o$. For example, to obtain the cost of picking up phone at the sofa, we sample 5 free locations around the sofa based on the occupancy grid map. For each low-level sample state $s^l_i$, we denote its trial cost as $\tilde{c}(s^l_i), i=1,...,N_l$. We use the average of all the trial costs as the empirical action cost $\tilde{c}(a_k, s^h_k)$ as follows:
\begin{gather}
    \tilde{c}(a_k, s^h_k)  = \frac{1}{N_l} \sum_{i=1}^{N_l} \tilde{c}(s^l_i).
\end{gather}

In order to reflect realistic action costs so that the LLM planner can better estimate unknown costs, we design the cost of each high-level action in a nuanced way as follows.

\textbf{Navigation Cost $c^{nav}$:}
When $a_k$ is \textit{navigate(room/furniture)}, the cost function $c(a_k, s^h_k)$ is instantiated as a navigation cost $c^{nav}(a_k, s^h_k)$. We explicitly design the cost as follows
\begin{equation}
    c^{nav} \triangleq \gamma^{nav}  cc^{nav} + t^{nav} + d^{nav} ,
\end{equation}
where $cc^{nav}$ is the collision count which denotes the number of robot colliding with a wall or a door and needs to replan navigation, $t^{nav}$ is the navigation time, $d^{nav}$ is the navigated distance, and $\gamma^{nav}$ is a normalizing factor of collision count since its value is much smaller than $t^{nav}$ and $d^{nav}$.

\textbf{Object Manipulation Cost $c^{man}$:}
When $a_k$ is \textit{pickup(object)} or \textit{place(object)} and $s^h_k$ is \textit{at(furniture)}, the cost function $c(a_k, s^h_k)$ is instantiated as an object manipulation cost $c^{man}(a_k, s^h_k)$. This cost category denotes the cost of robot manipulating an object or furniture in front of it.  We explicitly design the cost as follows
\begin{equation}
    c^{man} \triangleq \gamma^{man}  (1 - sr^{man}) + t^{man},
\end{equation}
where $sr^{man}$ is the success rate of picking up or placing an object (then $1-sr^{man}$ is the failure rate), $t^{man}$ is the pickup or place time, and $\gamma^{man}$ is a normalizing factor of failure rate since its value is much smaller than $t^{man}$.

% MCTS: sampling-based search, cost function is computed from simulated rollouts, needs heavy computation to simulate multiple rollouts.
% Since planning is essentially looking into the future, we need to estimate the action cost before the robot really executes it. 

%An intuitive example is that a robot tries to pick up a coffee cup by using its manipulator. The difficulty of executing the pick-up action should be uniformly distributed given the distance between the robot's arm and the coffee cup.

% The world model is difficult to model (doors, rooms, objects) and the robot dynamics is high-dimensional and nonlinear (mover, manipulator, sensor). Given the unknown dynamics of the robot operating in the local workspace, we need to estimate the cost of executing the action by a low-level motion planner, so that we can interleave it with the high-level LLM planner. 

\textbf{Update Known Action Costs $C^{known}$:}
Each time the motion planner executes a sequence of actions, it will collect empirical action costs in the real world. For \textit{navigate} action, the empirical cost is collected by A* on a grid map. For \textit{pick up} or \textit{place} action, the empirical cost is collected through the IK solution. We update these newly collected action costs in the known action costs $C^{known}$. If the newly collected action cost $\tilde{c}(a_k, s^h_k)$ has the same action-state pair $(a_k, s^h_k)$ as a known action cost $c^{known}(a_k, s^h_k) \in C^{known}$, we fuse them by taking average of the two cost values as a more accurate updated cost value:
\begin{gather}
    c^{updated}(a_k, s^h_k) = \frac{\tilde{c}(a_k, s^h_k) + c^{known}(a_k, s^h_k )}{2}\\
    C^{known} \leftarrow C^{known} \cup \{c^{updated}(a_k, s^h_k)\}
\end{gather}
If the newly collected action cost  $\tilde{c}(a_k, s^h_k)$ is never seen in $C^{known}$, we directly append it
$C^{known} \leftarrow C^{known} \cup \{\tilde{c}(a_k, s^h_k)\}$. In this way, as the motion planner collects more action costs, the LLM planner will be more accurately grounded with real-world physical details.

\subsection{Interleaving LLM Planner with Motion Planner through Multimodal Action Cost Similarity Function}

The primary motivation behind our interleaved planning is to enable the LLM planner to account for nuanced physical realities of a complex environment while guiding the low-level motion planner. Our key insight is that planning is essentially looking into the future; therefore, it is beneficial to estimate costs of potential actions before the robot really executes them. In this way, the estimated action costs help prune high-cost branches during the LLM planning process before the motion planner really executes these difficult actions. Specifically, we propose a multimodal action cost similarity function $\mathcal{F}^{ms}$ to estimate unknown navigation and manipulation costs, so that the LLM planner consistently incorporates action costs uploaded by the motion planner, thus achieving near-optimal plans as the mission is going on.

Assume at the current timestep the robot has already executed some actions and collected a set of real action costs $C^{known}$. We separate these known action costs into two parts  $C^{known} = \{C^{naved}, C^{maned} \} $: the known navigation costs $C^{naved}$ and the known manipulation costs $C^{maned}$. For each executed navigation action, $C^{naved}$ record its navigated path $p^{naved}$ and its navigation cost value $c^{naved}$. For each executed manipulation action, $C^{maned}$ record its executed action $a^{maned}$ and its manipulation cost value $c^{maned}$. After the LLM generates $M$ task plan candidates $\{\Pi_1, ..., \Pi_M \}$, we calculate the total estimated cost $\hat{c}^{\Pi_i}_{total}$ of each task plan candidate $\Pi_i = \{a_1,...,a_{N_i}\}, i=1,..,M$. Same as $C^{known}$, we separate the task plan into two parts $\Pi_i = \{ \Pi^{nav}_i, \Pi^{man}_i \}$: the navigation actions $ \Pi^{nav}_i $ and the manipulation actions $  \Pi^{man}_i$.

\textbf{Estimating Unknown Navigation Costs through Path Similarity Function:}
Different from the \textit{pickup} action associated with objects and furnitures, the \textit{navigate} action does not hold too much semantic information for the LLM to reasonably infer unknown action costs. Therefore, we propose to quantitatively compute the overlapping percentage of two paths so that we can better infer the unknown navigation costs from the known ones.

We calculate the overlapping percentage $P^o(p_i, p_j)$ between twp paths $p_i$ and $p_j$ as:
\begin{equation}
    P^o(p_i, p_j) \triangleq 100 * (1 - \frac{d^{i,j}_{am}}{\epsilon_d}) + 100 * (1 - \frac{d^{j,i}_{am}}{\epsilon_d} ),
    \label{equation path similarity}
\end{equation}
where $d^{i,j}_{am}$ is the average of the closest distances between each point in the path $p_i$ to any point in another path $p_j$, $d^{j,i}_{am}$ is the average of the closest distances between each point in the path $p_j$ to any point in another path $p_i$,  and $\epsilon_d$ is a hyperparameter distance to consider two points as overlapping, beyond which overlapping is 0\%. Through this symmetric definition, we comprehensively account for overlapping from both path directions and give an accurate sense of mutual proximity.

For each navigation action $ a^{nav}_k \in  \Pi^{nav}_i $, we estimate its unknown cost through the path similarity function \eqref{equation path similarity}. First, we extract the start furniture $fur^s_k$ and destination furniture $fur^d_k$  from the navigation action $a^{nav}_k$ and the current high-level state $s^{h,nav}_k$. Second, we use A* path planner to compute a \textit{presumed} path $p^{pre}_k$ from $fur^s_k$ to $fur^d_k$ based on the occupancy grid graph $\mathcal{G}^o$. We call this path \textit{presumed} because this path is pre-planned in advance and not yet executed by the robot in the real world. We do not know if this path will lead to robot collision or not. Last, we estimate the navigation cost  $\hat{c}(a^{nav}_k)$ by computing the path similarity between this presumed path $p^{pre}_k$ and all the navigated paths and navigation costs $(p^{naved}_i, c_i^{naved}) \in C^{naved}, i=1,...,N^{naved}$ as follows:
\begin{equation}
    \hat{c}(a^{nav}_k) = \sum^{N^{naved}}_{i=1} c_i^{naved} \cdot P^o(p^{pre}_k, p_i^{naved})
\end{equation}

\textbf{Estimating Unknown Manipulation Costs through Semantic Similarity Function:}
For each manipulation action $ a^{man}_k \in  \Pi^{man}_i $, we leverage LLM as a semantic similarity function to estimate its unknown action cost. First, we extract the corresponding object $obj^{man}_k$ and furniture $fur^{man}_k$ from the manipulation action  $ a^{man}_k$. We encode three semantic attributes $A^s$ of $(obj, fur)$ into the LLM prompt, which are \textit{\{location, category, usage\}}. We implement the same process for all the manipulated actions $a^{maned} \in C^{maned}$. Second, for all the manipulated action cost values $c^{maned} \in C^{maned}$, we convert its numerical value into textual space for the LLM to understand through an encoding function:
\begin{equation} 
f^{en}(c^{maned}) \triangleq \left\{  \begin{array}{ll} 
\text{hard},   &      c^{maned} > 15 \\
\text{medium},  &       5 \leq c^{maned} \leq 15  \\
\text{easy},    &       c^{maned} < 5 
\end{array} \right. 
\end{equation}
Then we ask the LLM to infer a textual cost $\hat{c}^{text}(a^{man}_k) $ of the action $a^{man}_k$ given all the manipulated actions and cost values $(a^{maned}_i, c_i^{maned}) \in C^{maned}, i=1,...,N^{maned}$ as follows:
\begin{gather}
\label{equation semantic similarity}
 (obj^{man}_k, fur^{man}_k) \leftarrow a^{man}_k\\
 (obj^{maned}_i, fur^{maned}_i) \leftarrow a^{maned}_i \\
 prompt \leftarrow A^s(obj^{man}_k, fur^{man}_k, obj^{maned}_i, fur^{maned}_i)\\
\hat{c}^{text}(a^{man}_k) \leftarrow LLM(prompt, a^{man}_k, f^{en}(c^{maned}_i))
\end{gather}
Note that the textual output from LLM can be \textit{unknown} since LLM finds it unreasonable to infer a manipulation cost without strong semantic similarity.

Likewise, we convert the textual cost $\hat{c}^{text}(a^{man}_k) $ into a numerical value through a decoding function:
\begin{equation} 
f^{de}(c^{text}) \triangleq \left\{  \begin{array}{ll} 
20,   &       \text{$c^{text}$ = hard} \\
10,  &         \text{$c^{text}$  = medium}  \\
5,  &          \text{$c^{text}$ = easy} \\
0,  &       \text{$c^{text}$  = unknown} 
\end{array} \right. 
\end{equation}

Finally we obtain the unknown manipulation cost  $\hat{c}(a^{man}_k)$ as follows:
\begin{gather}
\hat{c}(a^{man}_k) \leftarrow f^{de}(\hat{c}^{text}(a^{man}_k))
\end{gather}

%The large scene graph has a large number of elements. We can leverage generalized planning capabilities of LLMs. We could use LLM to generalize the action cost across different scenarios. The large-scale scene graph leads to a large search tree with high branching factors. Therefore, we introduce a branch prune method that leverages a lower bound on task completion costs to reduce the number of nodes that need to be considered in the search tree.

\textbf{Estimating Total Cost of Task Plan Candidates:}
After obtaining both the estimated navigation cost and the estimated manipulation cost, we calculate the total estimated cost of the task plan candidate $\Pi_i$  as follows:
\begin{equation}
    \hat{c}^{\Pi_i}_{total} = \sum^{N^{nav}}_{k=1} \hat{c}(a^{nav}_k)  +  \sum^{N^{man}}_{k=1} \hat{c}(a^{man}_k) * \frac{N^{man}_{valid}}{N^{man}},
\end{equation}
where $N^{nav}$ is the total number of estimated navigation costs,  $N^{man}_{valid}$ is the number of estimated manipulation costs which are non-zero, i.e.,  $\hat{c}^{text}(a^{man}_k) $ is not \textit{unknown}, and $N^{man}$ is the total number of estimated manipulation costs. Finally, we select the best task plan candidate $\Pi_*$ with minimum costs as follows:
\begin{gather}
    \Pi_* = \arg\min_{\hat{c}^{\Pi_i}_{total}}\Pi_i, i=1,...,M
\end{gather}

%%%%%%%%%%%%%%%%%%%%%%%%%%%%%%%%%%%%%%%%%%%%%%
\section{Experiments}
\label{experiments}

Latest works \cite{rana2023sayplan, honerkamp2024language} mainly guide the robot to fulfill a single human command in a  simple environment in a one-shot manner. We aim to go beyond by exploring three questions ``Can the robot consistently fulfill a bunch of human commands? Can the robot self-improve its performance given so many human commands? Can the overall performance maintain stable under so many uncertainties and constraints in a complex real world?'' As defined as our \textit{generalized multi-object collection} problem in Section \ref{problem formulation}, we ask robots to perform an open set of abstract human commands. Through extensive evaluation, we demonstrate that under the guidance of our interleaved planning algorithm, the robot can perform these commands with a good balance of quality and efficiency in the complicated environment.

\subsection{Experiment Setup}
% how to design a specific experiment to showcase the advantage of our algorithm?

We consider a long-horizon mission consisting of multiple human commands. Image a family get up in a workday morning and each family member issues commands to the household robot for help in a short time window.
\begin{itemize}
    \item Command 1 from mother ``My son needs to have the breakfast. Set it up on the dinning table.''
    \item Command 2 from son ``I have an online meeting in 10 minutes. Set it up in my bedroom.''
    \item Command 3 from father ``I need to read a book for a break. Also, let me check what is going on in USA today.''
\end{itemize} 
The robot is a ManipulaTHOR mobile manipulator with a 6-DoF grasping arm and an on-boarding camera \cite{ehsani2021manipulathor}. We use A* to execute the \textit{navigate} action in the occupancy grid graph $\mathcal{G}^o$. For \textit{pickup/place} actions, we execute object manipulation through IK solutions when the robot arrives at the local area of a furniture and turns toward the \textit{object} through camera detection. We select an extremely large household environment in the ProcTHOR simulator \cite{procthor}, which has 9 rooms, 26 furnitures, 30 objects. All the hyperparameter values are shown in Table \ref{table hyperparameters}.

\begin{table}
\caption{Hyperparameter values in our experiments.}
\vspace{-2mm}
\begin{center}
\renewcommand{\arraystretch}{1.2}
\begin{tabular}{| c || c  |}
    \hline
    \textbf{Hyperparameters } & \textbf{Value} \\
    \hline

    number of task plan candidates $M$ &  3    \\
   
    \hline 
 
   LLM temperature parameter $\sigma$ &  0.8 \\

    \hline

    collision count normalizing factor $\gamma^n$   & 10   \\

     \hline

    success rate normalizing factor $\gamma^m$ & 100  \\

    \hline 
    
    object fulfillment rate normalizing factor $\gamma^o$  & 100  \\

  \hline
\end{tabular}
\label{table hyperparameters}
\end{center}
\vspace{-5mm}
\end{table}

\textbf{Metrics:}
Consistent with our problem formulation which considers robotic action costs, we directly use all these costs as straightforward metrics to evaluate the mission performance: navigation collision counts $cc^{nav}$, navigated distance $d^{nav}$, manipulation success rate $sr^{man}$, total execution time $t^{exe} = t^{nav} + t^{man}$ consisting of both navigation time $t^{nav}$ and manipulation time $t^{man}$. Since in our problem formulation fulfilling multiple human commands requires the robot to collect multiple objects, we introduce another metric $sr^{obj}$ which denotes the object fulfillment rate. Last, we define an overall metric $m_{overall}$ to consider all the action execution and command fulfillment progress together as overall performance evaluation. 
\begin{align}
  m_{overall} & \triangleq  \gamma^{nav} cc^{nav} + t^{exe}  + d^{nav} \\
  & + \gamma^{man} (1 - sr^{man}) + \gamma^{obj}(1 - sr^{obj}),
\end{align}
where $\gamma^{obj}$ is a normalizing factor of object fulfillment rate.

%We manually pre-construct a scene graph as the representation of this large-scale environment. Motivated by \cite{honerkamp2024language}, we maintain a structured representation of the scene graph, which is better to ground LLMs for less hallucination. 

% ToolBench \cite{xu2023tool}: Task: Move the robot to {position}, and {move} the arm by {height}. API Calls: robot.move\_to({x}, {y}) robot.raise\_arm({h}) Random Value Pool position: location (20, 30) move: raise height: 5cm x: 20 y: 30 h: 5 Training Data Task: Move the robot to location (20, 30), and raise the arm by 5cm. API Calls: robot.move\_to(20, 30) robot.raise\_arm(5)position: the corner move: raise height: 0.5 meter x: 0 y: 0 h: 50 ... position: (0, 40) move: drop height: 10cm x: 0 y: 40 h: -10

\subsection{Preliminary Results of Multimodal Similarity Function}

The most significant prerequisite for our Inter-LLM algorithm to work is the multimodal action cost similarity function. We present the following preliminary results of this key design. 

\begin{figure*}
\centering

\includegraphics[width=0.3\textwidth]{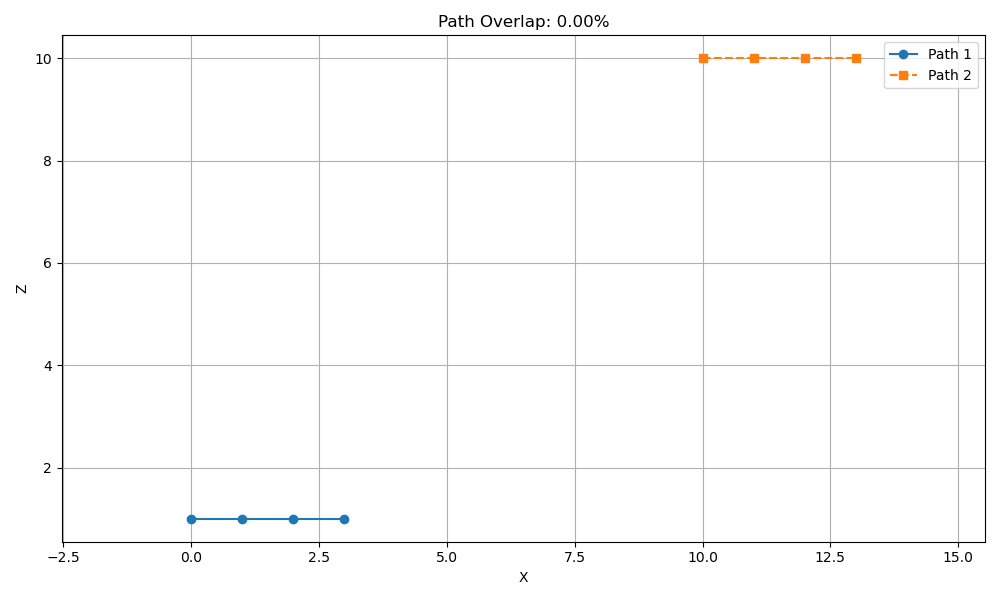}
\includegraphics[width=0.3\textwidth]{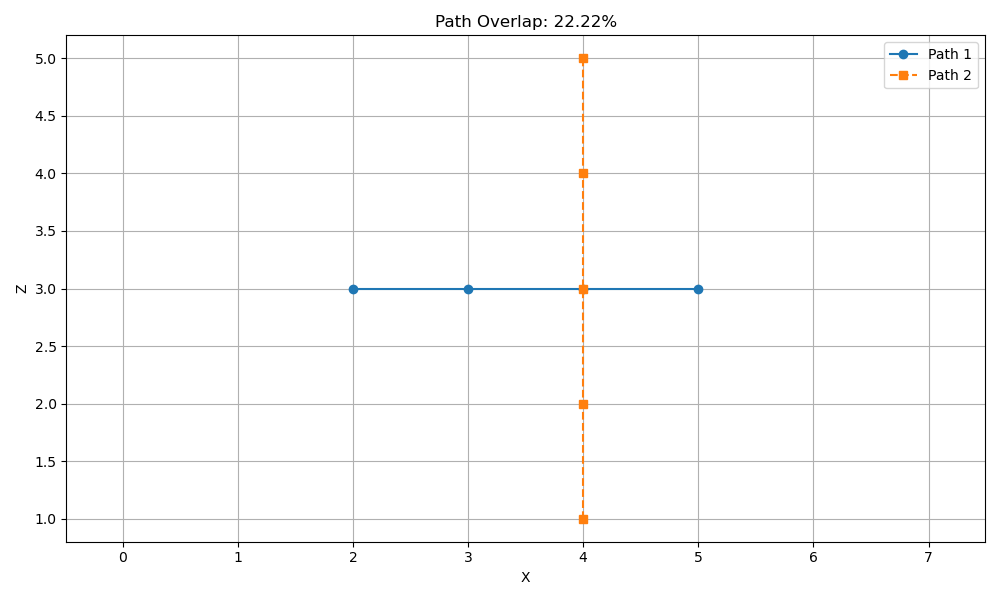}
\includegraphics[width=0.3\textwidth]{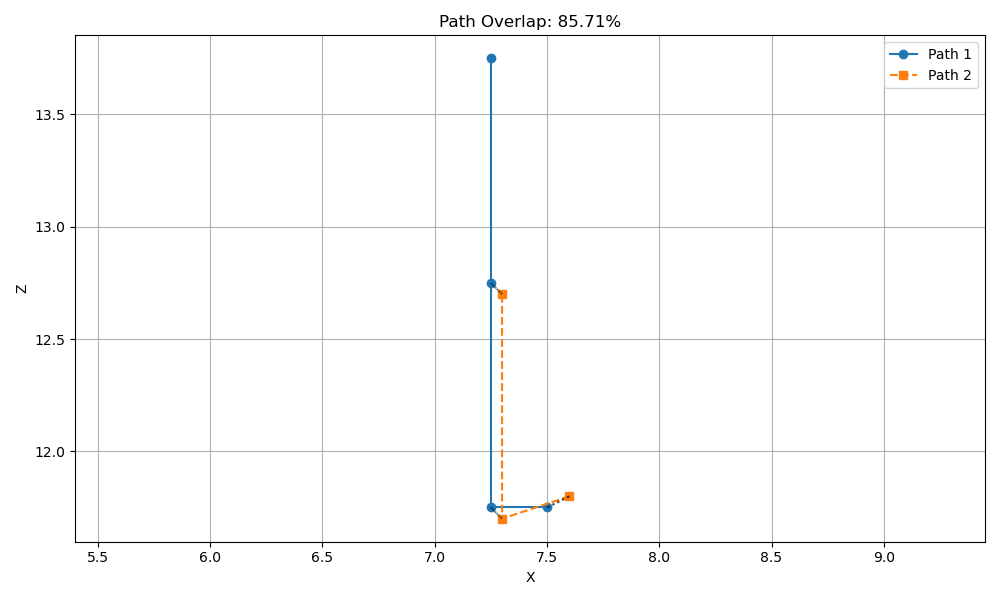}

\caption{Preliminary results of path similarity. The more overlapped the two paths are, the higher the path similarity percentage is.}
\label{figure path_similarity}
\end{figure*}

\textbf{Path Similarity Results:}
As shown in Figure \ref{figure path_similarity}, we demonstrate the effectiveness of our path similarity function. It is expected that the more overlapped the two paths are, the higher the path similarity percentage is. In this way, the path similarity function can help the LLM planner circumvent paths which will lead to collision by comprehensively considering all the navigated paths and their corresponding costs.

\textbf{Semantic Similarity Results:}
The quality of inferring semantic similarity between manipulation action costs is decided by the temperature hyperparameter $\sigma$ of LLM. We present the results in Table \ref{table result_llm_temperature_parameter} to evaluate how LLM relies on it for unseen cost generalization. For example, assume we obtain the known cost of \textit{pickup(phone), at(table\_3)} is \textit{easy}, and the LLM needs to infer the action cost \textit{pickup(remote\_control), at(table\_3)}. If we set $\sigma = 1.0$ (strict reasoning), the LLM will say ``I am not sure if \textit{remote\_control} is similar to \textit{phone} at this table, so I can’t infer the action cost \textit{pickup(remote\_control), at table\_3}.'' If we set $\sigma = 0.2$ (loose reasoning), the LLM will say ``\textit{pickup(remote\_control), at(table\_3)} should be similar to \textit{pickup(phone), at(table\_3)}, since they are both \textit{pickup} actions.''

\begin{table}
\caption{Results of LLM temperature hyperparameter $\sigma$ on inferring semantic similarity between manipulation action costs.}
\vspace{-2mm}
\begin{center}
\renewcommand{\arraystretch}{1.0}
\begin{tabular}{| c || c  |}
    \hline
    \textbf{Temperature Hyperparameter $\sigma$} & \textbf{Semantic Similarity Accuracy} \\
    \hline
     0.0 &  11\%    \\
   
    \hline 
 
   0.2  &  20\%  \\

    \hline

   0.4  & 35\%   \\

     \hline

   0.6 & 57\%  \\

     \hline

   \textbf{0.8}  & \textbf{73\%}   \\

     \hline

   1.0  & 66\%   \\

  \hline
\end{tabular}
\label{table result_llm_temperature_parameter}
\end{center}
\vspace{-5mm}
\end{table}

\subsection{Full Results with Baseline Comparison}

\begin{figure}
        \centerline{\includegraphics[width=0.5\textwidth]{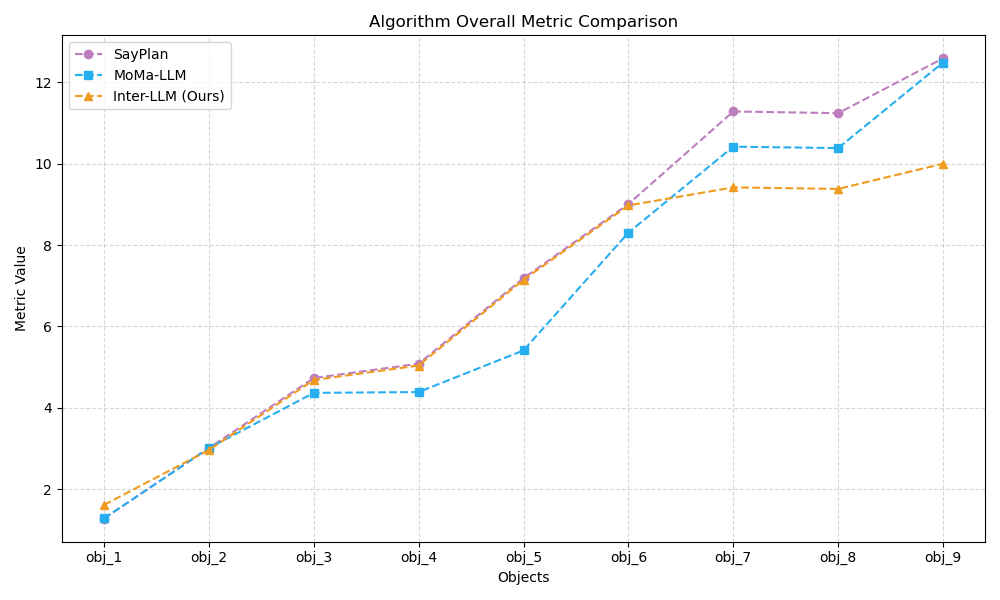}}
        \caption{Results of algorithm comparison by overall cost metric. We evaluate our Inter-LLM algorithm and the two baselines SayPlan \cite{rana2023sayplan} and MoMa-LLM \cite{honerkamp2024language} in the whole mission which requires navigating towards and manipulating 9 objects.}
        \label{result_overall_metric}
\end{figure}

\begin{table*}
\caption{Comprehensive metric results of algorithm comparison between SayPlan \cite{rana2023sayplan}, MoMa-LLM \cite{honerkamp2024language}, and our Inter-LLM.}
\vspace{-2mm}
\begin{center}
\renewcommand{\arraystretch}{1.2}
\begin{tabular}{| c | c || c | c | c | c | c | c | c | c | c|}
    \hline

    \multicolumn{2}{|c||}{\multirow{2}{*}{\textbf{Algorithms}}} &  \multicolumn{3}{c|}{\textbf{command\_1}} & \multicolumn{3}{c|}{\textbf{command\_2}} & \multicolumn{3}{c|}{\textbf{command\_3}} \\

       \cline{3-11}

     \multicolumn{2}{|c||}{}  & \textbf{obj\_1} & \textbf{obj\_2} & \textbf{obj\_3} & \textbf{obj\_4} & \textbf{obj\_5} & \textbf{obj\_6} & \textbf{obj\_7} & \textbf{obj\_8}  & \textbf{obj\_9} \\

    \hline

  \multirow{3}{*}{Navigation Collision Counts} &  SayPlan  &  0 &  1 & 1 & 1 & 3 & 0 & 2 & 0 & 0 \\

    &  MoMa-LLM  &  0 & 1 & 1 & 1 & 3 & 4 & 3 & 0 & 2  \\

   & Inter-LLM (Ours)  &  1 & 0 & 1 & 1 & 3 & 0 & 2 & 0  & 1 \\

   \cline{1-11}

    \multirow{3}{*}{Navigated Distance (m)} & SayPlan  & 10.0 & 28.0 & 28.2  & 31.0 & 31.5 & 11.0 & 31.0 & 8.0 & 11.2  \\

    &  MoMa-LLM  &  10.0 & 28.0 & 56.5 & 31.0 & 40.7 & 71.5 & 66.5 & 8.0 & 52.5  \\

   & Inter-LLM (Ours)  &  20.0 & 15.0 & 28.2 & 31.0 & 31.5 & 11.0 & 27.7 & 8.0 & 43.5 \\

    \cline{1-11}

    \multirow{3}{*}{Manipulation Success Rate} & SayPlan  & 0\% & 0\% & 0\% & 40\% & 0\% & 0\% & 0\% & 30\% & 0\%  \\

    &  MoMa-LLM  &  0\% & 0\% & 10\% & 60\% & 20\% & 5\% & 6\% & 30\% & 4\%  \\

   & Inter-LLM (Ours) & 0\% & 0\% & 0\% & 40\% & 0\% & 0\% & 40\% & 30\% & 30\% \\

   \cline{1-11}

    \multirow{3}{*}{Total Execution Time (s)} & SayPlan  & 17.5 & 35.2 & 34.0 & 34.0 & 48.7 & 71.2 & 76.6 & 17.8 & 23.5 \\

    &  MoMa-LLM  &  17.8 & 34.9 & 79.9 & 27.6 & 56.6 & 183.0 & 121.7 & 18.5 &  142.5 \\

   & Inter-LLM (Ours)  & 31.0 & 20.1 & 33.8 & 34.5 & 49.1 & 71.8 & 34.1 & 18.2 & 38.5 \\

     \cline{1-11}

    \multirow{3}{*}{Object Fulfillment Rate} & SayPlan  & $\times$ & $\times$ & $\times$ & $\surd$ & $\times$ & $\times$ & $\times$ & $\surd$ & $\times$ \\

    &  MoMa-LLM  &  $\times$ & $\times$ & $\surd$ & $\surd$ & $\surd$ & $\surd$ & $\surd$ & $\surd$ & $\surd$ \\

   & Inter-LLM (Ours)  & $\times$ & $\times$ & $\times$ & $\surd$ & $\times$ & $\times$ & $\surd$ & $\surd$ & $\surd$  \\

  \hline

\end{tabular}

\label{table result_all_metrics}

\end{center}
\vspace{-6mm}
\end{table*}

%\begin{figure*}
 %      \centerline{\includegraphics[width=\textwidth, height=5cm]{experiments/result_by_commands.png}}
  %      \caption{Results of algorithm comparison by human commands. We evaluate our Inter-LLM algorithm and the two baselines SayPlan and MoMa-LLM in the whole mission consisting of three human commands.}
   %     \label{figure result_by_commands}
%\end{figure*}

We present comprehensive results of running the three algorithms in the ProcThor scene train\_1. In total, we evaluate ten extremely long missions, where each mission consists of three human commands. The robot needs to plan a complicated sequence of up to 24 \textit{navigate}, \textit{pickup}, \textit{place} actions, which are related to 8 rooms, 12 furnitures, and 9 objects in average. We compare with two latest works SayPlan \cite{rana2023sayplan} and MoMa-LLM \cite{honerkamp2024language}. For a fair comparison with the baselines, we use the same LLM prompt as our Inter-LLM algorithm to generate the initial task plans, but of course, the subsequent planning process is determined by the baseline algorithms themselves. For instance, MoMa-LLM keeps replanning whenever one single action fails, while our Inter-LLM leverages the multimodal similarity function to improve the whole mission performance in a systematic manner. Also, we make sure the initial task plans generated by all the algorithms have the same number of objects to manipulate. Please see the full video \url{https://youtu.be/C3CaJSHZFes}

We compare overall algorithm performance by the overall metric $m_{overall}$. As shown in Figure \ref{result_overall_metric}, our algorithm Inter-LLM maintains stable performance even as the number of objects — thus the mission's uncertainty and complexity — increase. In contrast, the baselines SayPlan and MoMa-LLM struggle to improve themselves over this long process. It is worth noting that MoMa-LLM outperforms Inter-LLM early in the mission (up to obj\_6), but its performance degrades as mission complexity grows, which highlights its limited ability to adapt to increasing uncertainty. SayPlan consistently performs the worst, as it follows an open-loop LLM-motion planning pipeline that relies solely on semantic heuristics, without accounting for realistic action costs. Overall, Inter-LLM improves the mission performance by 30\% compared to the baselines, producing near-optimal plans.

For a more detailed comparison, we evaluate the algorithms using fine-grained metrics that reflect real-world execution costs: navigation collision counts $cc^{nav}$, navigated distance $d^{nav}$, manipulation success rate $sr^{man}$, total execution time $t^{exe}$, and the object fulfillment rate $sr^{obj}$. As shown in Table \ref{table result_all_metrics}, for command\_2, MoMa-LLM successfully retrieves all three objects, but at the cost of significantly longer execution time, higher navigated distance, and more navigation collisions compared to both SayPlan and our Inter-LLM. Although Inter-LLM retrieves only obj\_4 for command\_2, it self-improves a lot when trying to fulfill command\_3 and successfully retrieves all three objects. Eventually, it achieves the shortest total execution time, moderate collision count, and the highest manipulation success rate, demonstrating a strong balance between efficiency and quality.

%Random: collect multiple objects randomly

%Greedy: Additionally, Greedy was much faster than our method in planning time yet lead to lower success rate within the time budget and longer total time. Our intuition is that, while Greedy prioritizes looking at a location with the highest belief, our algorithm considers the optimal search of multiple objects in a sequence.

%Oracle: human user controls the robot to collect multiple objects through keyboard

%ConceptGraphs: Open-Vocabulary 3D Scene Graphs for Perception and Planning \cite{gu2023conceptgraphs}

% Optimal Scene Graph Planning with Large Language Model Guidance \cite{dai2023optimal}

% SayPlan \cite{rana2023sayplan}: their action cost is essentially the predicates in the task planner, not the physical cost at the motion planner. For example, SayPlan considers that a \textit{pickup(banana)} action might fail if the predicate of the robot is ``holding something'', but it doesn't consider the real cost of the manipulator grasping the banana.

\subsection{Analysis}

SayPlan consistently performs the worst across all metrics. This is because it follows an open-loop LLM-motion planning pipeline that relies solely on semantic heuristics, without considering real-world action costs. As a result, it struggles to deal with the high uncertainty and complexity introduced by long-horizon missions and large environments.

We identify two main reasons why MoMa-LLM performs worse than Inter-LLM: 1) MoMa-LLM allows the robot to open doors—treated as obstacles in our formulation—so all navigate actions appear successful. However, it does not record any meaningful information of the navigation process (e.g., door collisions) that could inform and improve future LLM planning.  In contrast, our Inter-LLM leverages the path similarity function to help the LLM  planner consider historical navigation costs for improving navigation performance over time; 2) MoMa-LLM considers ``success or failure'' of manipulation actions and keeps replanning until success. However, it only responds after execution fails — essentially a trial-and-error process without look-ahead planning. This can be highly inefficient especially for difficult actions. Inter-LLM, by contrast, proactively estimates the difficulty of each action using a multimodal similarity function, pruning hard-to-execute steps in advance and saving significant time.

Through experiments, we observe that Inter-LLM and MoMa-LLM often produce similar initial task plans based on the same LLM prompt, selecting the same object types but at different furniture locations. For example, for the command\_3 ``I need to read a book for a break. Also, let me check what is going on in USA today.'', both Inter-LLM and MoMa-LLM generate an initial task plan of picking up book, newspaper, and cellphone. However, Inter-LLM uses the multimodal similarity function $\mathcal{F}^{ms}$ to evaluate plan costs and choose the easiest navigate and pickup actions, while MoMa-LLM navigates to a random furniture for hard pickup. If MoMa-LLM picks a hard-to-reach object, it enters a trial-and-error replanning loop, which can exceed its budget without fulfilling the human command. More critically, this gap widens over long missions and complex environments. Our multimodal similarity function continually refines its cost estimation using feedback from motion planner, enabling increasingly optimal task plans. However, MoMa-LLM lacks this forward-looking mechanism and relies on reactive replanning after failures, making it unscalable with mission complexity and uncertainty.

\section{Conclusion}
\label{conclusion}

We propose a novel interleaved LLM and motion planning algorithm for generalized object collection in large scene graphs. Compared with latest works, our algorithm achieves a strong balance of quality and efficiency through LLM semantic heuristics, symbolic feasibility checking, and high-cost action pruning by the multimodal action cost similarity function. Future works could be: 1) Introduce task scheduling before planning to efficiently partition long-horizon missions, allowing simultaneous processing of human commands instead of one by one; 2) Extend planning on pre-built scene graphs to POMDP planning on unknown ones; 3) Enhance the similarity function with visual feedback (e.g., furniture layout images during manipulation) to enable more accurate action cost estimation.

\bibliography{IEEEabrv, reference.bib}
\bibliographystyle{IEEEtran}

\end{document}